\documentclass[12pt]{elsarticle}
\usepackage[english]{babel}
\usepackage{amsmath} 
\usepackage{amssymb}
\usepackage{array}
\usepackage{graphicx}
\usepackage[labelformat=simple]{subcaption}
\usepackage{fancyhdr}
\usepackage{booktabs,multicol,multirow}
\usepackage{xcolor}
\usepackage{afterpage}
\usepackage[colorinlistoftodos,textsize=small]{todonotes}
\usepackage{url}
\usepackage{hyperref}
\usepackage{longtable}

\newcommand{\cx}{\mathbf{x}}

\makeatletter
\def\ps@pprintTitle{%
 \let\@oddhead\@empty
 \let\@evenhead\@empty
 \def\@oddfoot{}%
 \let\@evenfoot\@oddfoot}
\makeatother

\begin{document}

\begin{frontmatter}


\title{DeepRadiologyNet: Radiologist Level Pathology Detection in CT Head Images
% jtm: footnote not showing up, info was put in acknowledgements.
%\thanks{Work conducted in 2015 and 2016 while SS, ST, AV were supported in part by DeepRadiology, INC. Disclosure covered by US 62/275,064; January 5, 2016.}
}


\author[dr]{J.~Merkow}
% \ead[WebPage.com]{jameson@deepradiology.com}
\author[dr]{R.~Lufkin}
\author[dr]{K.~Nguyen}
\author[ucla]{S.~Soatto}
\author[ucsd]{Z.~Tu}
\author[oxford]{A.~Vedaldi}

\address[dr]{DeepRadiology, INC.}
\address[ucla]{University of California, Los Angeles}
\address[ucsd]{University of California, San Diego}
\address[oxford]{University of Oxford}

\begin{abstract}
We describe a system to automatically filter clinically significant findings from computerized tomography (CT) head scans, operating at performance levels exceeding that of practicing radiologists. Our system, named DeepRadiologyNet, 
builds on top of deep convolutional neural networks (CNNs) trained using approximately 3.5 million CT head images gathered from over 24,000 studies taken from January 1, 2015 to August 31, 2015 and January 1, 2016 to April 30 2016 in over 80 clinical sites.
For our initial system, we identified {30} phenomenological traits to be recognized in the CT scans. To test the system, we designed a clinical trial using over 4.8 million CT head images (29,925 studies), completely disjoint from the training and validation set, interpreted by {35} US Board Certified radiologists with specialized CT head experience. We measured clinically significant error rates to ascertain whether the performance of DeepRadiologyNet was comparable to or better than that of US Board Certified radiologists. DeepRadiologyNet achieved a clinically significant miss rate of 0.0367\% on automatically selected high-confidence studies.
Thus, DeepRadiologyNet enables significant reduction in the workload of human radiologists by automatically filtering studies and reporting on the high-confidence ones at an operating point well below the literal error rate for US Board Certified radiologists, estimated at 0.82\%.

\end{abstract}

%\begin{keyword} 
%Convolutional Neural Networks \sep 




%\end{keyword}

\end{frontmatter}



\section{Introduction}\label{sec:intro}

% Motivation
Analysis of medical imaging data is often one of the first steps in the diagnosis and determination of course of treatment, which in some cases must be determined within a few minutes, making the time spent by the radiologist on analysis a critical bottleneck. Assisted or automated analysis can help reduce the time necessary to arrive at a diagnosis. In addition, human error during routine diagnosis is often unavoidable even for highly trained medical professionals, for example due to human fatigue, inattention and distraction. Nevertheless, such errors can harm patients and drive up the cost of medical care, which has an adverse effect on the health care system at large. Assisted or automated analysis can help reduce the error in diagnosis. It can also make high-quality medical care possible in situations where no highly trained physicians are available, or where the cost of their services would be prohibitive. In fact, the use of machine learning techniques allows the benefits of training and experience to be shared globally by all systems, rather than each individual system being trained in isolation. 

% Define aim/scope
In this manuscript, we focus on computerized tomography (CT) as a  representative imaging modality, and on the detection of clinical pathologies such as intra-cranial bleeds as a  representative task. In particular we consider CT studies of the head (CT head), where a study is a collection of imaging data captured from the same subject during the same session, for instance a collection of a few tens to hundreds of two-dimensional (2D) slices comprising a volume image. Our aim is to build a system that can generate reports automatically for a large fraction of  CT head cases, while studies that our system does not generate reports for are referred to a human radiologist. The goal of this system is not to replace the radiologist entirely, but to reduce  human workload. 
The network identifies studies where it can generate a report with sufficient confidence, referring other cases, which may or may not be clinically significant, to a radiologist.
In this paper, we measure the \emph{clinically significant miss rate} within the studies that are not referred to a radiologist. Intuitively, our system can be made arbitrarily safe simply by reducing the number of cases that are reported on to zero.
While safe, this would not be useful. As we have progressed, the percentage of studies on which the network can report for a better-than-human error rate has steadily increased, thus making the approach viable to reduce the workload of human radiologists. For example, at literal error rates, DeepRadiologyNet reduces the load on human radiologists on over 40\% of studies.
In practice, we choose an operating point conservatively, with significantly lower error than the literal error rate of US Board Certified radiologists.

% Summary of contributions
Our system leverages on recent developments in Deep Learning.
However, off-the-shelf systems are insufficient to address our challenge. Hence, we make three key contributions: First, we train our neural networks from millions of CT head images professionally annotated, thus distilling them from the observation of thousands of hours of human labor. Second, we redesign state-of-the-art neural network architectures to better match the statistics of CT images, which differ substantially from the natural (everyday) images for which typical architectures are optimized. Third, we carefully evaluate the reliability of the resulting system and show that it can be used to identify, automatically, a large fraction of pathology with comparable or better overall accuracy than expert radiologists.

% Contribution 1: data
One attractive aspect of using Machine Learning to develop diagnostic systems is that, while the human visual system has evolved over millions of years to be attuned to interpret natural images, it is not naturally suited to interpret medical images. This is why training radiologists is a long process, and the mapping from non-optical sensory signals, as in CT or magnetic resonance (MR), to images that can be viewed by a human may entail information loss. This mapping is not necessary for an automated system, that can be trained to perform inference directly from raw sensory input, without the need for optical visualization. This offers the potential for pre-clinical diagnosis, before disease is manifest in an optical image rendered to a radiologist.

\begin{figure}[htp]
\includegraphics[width=\textwidth]{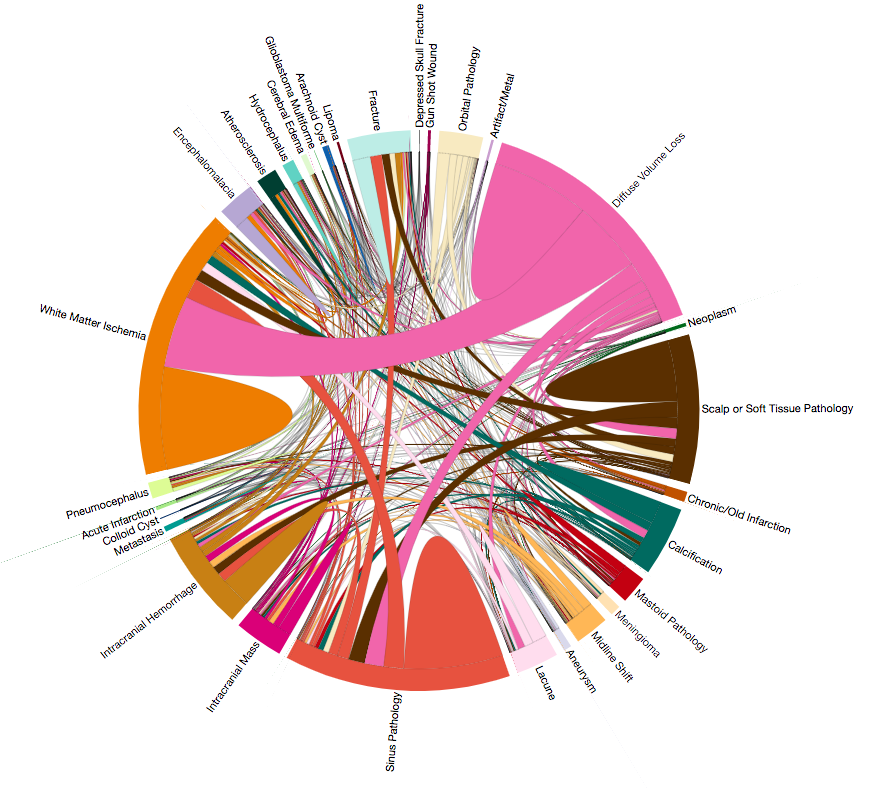}
\caption{Chord Diagram showing the co-occurrence of phenomenological traits within our trial dataset of 29,965 CT studies of the head, consisting of over 4.8 million images.
\label{fig:chord}
}
\end{figure}

These benefits can only materialize if neural networks can be trained from a sufficient quantity of high-quality data, the acquiring of which is often one of the most significant practical hurdles. In this work, we leverage a {large} curated collection of imaging studies to train a deep neural network, which is a parametric class of functions whose parameters can be adapted to fit a complex map from imaging data to classification outcomes matching that of expert radiologists. 

% Contribution 2: architectures
Detection comprises a binary classification task, as to whether a pathology is present, a multi-class classification (which of a set of pathologies), and localization (where in the volume is the pathology manifest).

Within Machine Learning, the use of deep neural networks for interpreting imaging data resurged to prominence in 2013 \cite{alexnet}, despite the key tools being available since long before \cite{rumelhart1986learning}, following the availability of large annotated datasets of natural images \cite{imagenet} as well as computing hardware initially developed for graphics rendering. Our initial attempt, in 2014, to exploit a network  pre-trained on ImageNet and fine-tuned to a relatively small number of imaging studies gave encouraging but far from human-level results due to the significantly different phenomenology. For example, natural image classification is unaffected by changes of intensity value so long as local ordering is not affected (contrast changes), whereas the intensity value recorded at a pixel of a CT scan, measured in Hounsfield units, is informative of certain classes of pathology. Humans cannot perceive absolute luminance, and their perception is largely contrast-invariant. Similarly, natural images are subject to visibility artifacts (occlusions), whereas medical imaging sensors are designed precisely to overcome occlusion. Large shape variations induced in the image domain by changes of vantage point in a natural image do not change the identity of the object being portrayed, whereas deformation of anatomical structures in a medical image are often indicative of pathologies.

In some respect, therefore, medical images are simpler than natural images, as the most detrimental sources of nuisance variability (viewpoint, illumination and partial occlusion) are absent. On the other hand, they are challenging in that subtle class-specific variations are often obfuscated by significant intra-individual variability. Whereas much of the effort in training classifiers for natural images goes to discard nuisance variability, most of the effort in training deep neural networks, and specifically convolutional ones (CNNs), goes to disentangling subtle class-specific variability from large intra-individual nuisance variability. The practical consequence of this is that simply downloading a pre-trained network and hoping that fine-tuning it on a small number of annotated medical images will achieve satisfactory performance is wishful thinking, and training from scratch in a modality-specific manner is necessary.

% Contribution 3: evaluation
In this manuscript, we describe a system, first disclosed at Radiological Society of North American (RSNA) in December 2016, that performs recognition of {30} traits in CT head images which has been developed over the course of multiple years.  This system, DeepRadiologyNet, is evaluated on a dataset that is orders of magnitude larger than previous works (over 4.8 million images) and shown to have a lower clinically significant miss rate than an estimated literary miss rate.

\section{Related Work}

Related work exploiting deep neural networks in medical imaging includes \cite{esteva2017dermatologist}, where dermatology images are automatically evaluated by the Inception V3 network trained and evaluated using nine-fold cross validation on a set of 129,450 images. The challenges in dermatology are different than in CT, and more akin to natural images, where there is irradiance variability due to the interplay of the reflectance and diffusion property of the tissues with the properties of the illuminant. In CT, the probing signal is not unstructured electromagnetic in the visible spectrum, but rather penetrating radiation in the X-ray band, that is undeflected by the tissues; furthermore, the data is volumetric, and the phenomenology is substantially different, measuring absorption (not subject to occlusion), rather than reflectance. 

Work on X-ray includes the recently disclosed CheXNet, that is claimed to reach radiologist-level pneumonia detection on chest x-rays with Deep Learning \cite{rajpurkar2017chexnet}. The validation set there is limited to 420 images, in our view insufficient to determine suitable performance with a sufficient level of confidence. As comparison, we test our system on a validation set of millions of images.

Additional modalities where deep learning has been deployed include fundus imaging to assess diabetic retinopathy \citep{gulshan2016development}, electrocardiogram for arrhythmia detection \citep{rajpurkar2017cardiologist}, and 
hemorrhage identification \citep{grewal2017radnet}.
Automated diagnosis from chest radiographs has received increasing attention with algorithms for 
pulmonary tuberculosis classification \citep{lakhani2017deep} and lung nodule detection \citep{huang2017added}. \cite{islam2017abnormality} studied the performance of various convolutional architectures on multiple datasets.
Recently, \cite{wang2017chestx} released a new large scale dataset ChestX-ray-14, with performance benchmarked using ImageNet pre-trained architectures.
Competition on this dataset has already begun with multiple works showing improved performance \citep{yao2017learning,rajpurkar2017chexnet}.
 
In general, the use of deep learning for medical imaging has been the subject of intense interest, including dedicated sessions at medical imaging conferences, and book publications. This is understandable, but the devil is in the detail, and we find the quality of the data, as well as its curation including the ontology of classes to be trained for, to be as critical as the choice of architecture or optimization scheme.

\section{Formalization}\label{sec:formalization}

 \begin{figure}[htp]
 \includegraphics[width=\textwidth]{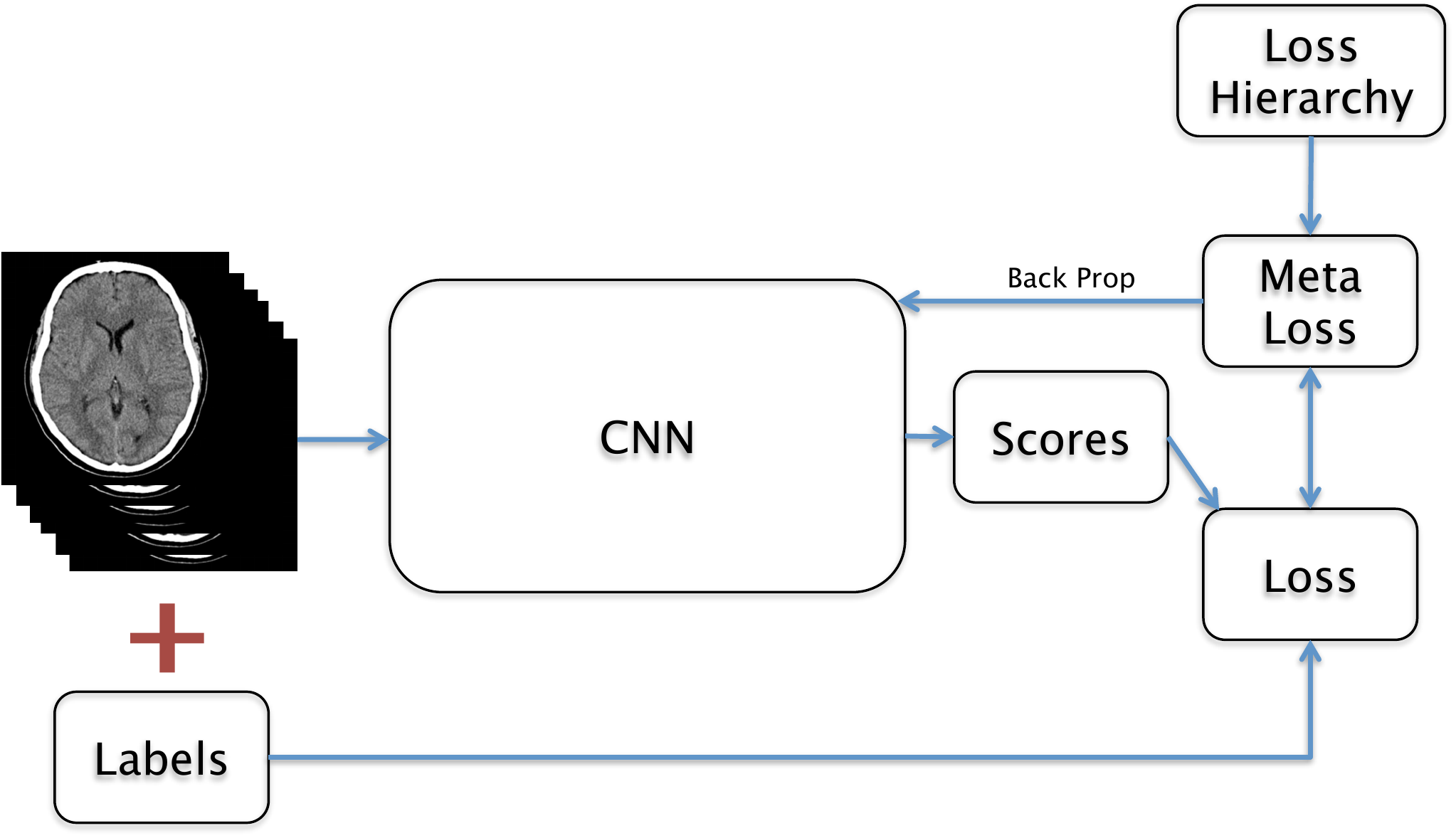}
 \caption{Illustration of hierarchal loss used to train DeepRadiologyNet.
 \label{fig:train}
 }
 \end{figure}

If we call $\cx = \{x_i, \ i = 1, \dots, N\}$ a set of images in a study, with $x_i \in [0, 1, 2, \dots, L-1]^{m\times n}$ an image with $L$ gray scale values, and $m\times n$ pixels, and $\cx^j$ one of $j = 1, \dots, M$ studies, and $y \in \{0, 1, \dots, K\}$ one of $K$ classes, each representing a disease or a specific phenomenological aspect of the study that might be of interest to a radiologist, then we are interested in constructing a classifier, $f: X \rightarrow Y; \ \cx\mapsto y$ that, in response to an input image or study, produces a label $y$. More in particular, we are interested in this function to be (not necessarily uniquely) determined by a set of parameters $w \in {\mathbb R}^P$, where $P$ can easily be in the tens of million. Furthermore, we want $f$ to be written as a simple and constant function (called a classifier) of a more complex function (called discriminant, or embedding, or representation) $\phi: X \times W \rightarrow {\mathbb R}_+^K; (x,w) \mapsto \phi_w(x)$ where $\phi_w(x)[k]$ denotes the $k$-th component of the $K$-dimensional vector $\phi_w(x)$. Assuming that medical images $x$ and their associated label $y$ are drawn from an unknown probability distribution $P(x, y)$, the optimal (Bayesian) discriminant would be the function
\begin{equation}
\phi_w(x)[k] = P(y = k | x)
\end{equation}
where the right-hand side is the posterior probability of the label given the image $x$. From now on we do not distinguish between images and studies, which is an application-dependent choice. In this work, we choose the discriminant $\phi_w(\cdot)$ among the class of functions represented by deep (multi-layer) neural networks. These are universal approximants \cite{cybenko1989approximation}, meaning that, given sufficient parameters, they can approximate any finite-complexity function. Machine learning-theoretic considerations are beyond the scope of this paper, where we simply assume that the optimal discriminant is within the chosen function class. 

If the optimal discriminant is available, inference proceeds simply by computing the maximum over all $k \in \{0, \dots, K\}$:
\begin{equation}
f(x) = \arg\max_k \phi_w(x)[k].
\label{eq:discriminant}
\end{equation}
The goal of learning is, given samples from the joint distribution $P(x, y)$, to determine the parameters $w$ so that, on average, the error made in approximating $f(x)$ with the right-hand side of \eqref{eq:discriminant} is smallest. The crux of the matter is that we cannot computer the average (expected value) with respect to $P(x,y)$ since we do not have access to it, so it is standard practice in Machine Learning to minimize the sample average (empirical loss), 
\begin{equation}
\hat w = \operatornamewithlimits{argmin}_w \sum_{(x_i, y_i) \sim P(x, y)} \ell_w(x_i, y_i)
\end{equation}
where $\ell(x,y)$ is the loss incurred when rendering the decision $y$ in response to the image $x$. Relating the empirical loss to the expected loss requires some assumptions on the distribution; studying this relation is the main subject of statistical learning theory, which we do not delve into here. Suffice for us to say that some form of regularization is typically necessary to ensure that the minimizer of the empirical loss bears some resemblance to the minimizer of the expected loss, and therefore can {\em generalize} to unseen samples \cite{vapnik1998statistical}. In the case of training convolutional neural networks (CNNs) using stochastic gradient descent (SGD), such regularization takes many forms, some implicit in the nature of SGD, some explicit (e.g., Dropout \cite{dropout}), others rooted in the choice of architecture (e.g., pooling). This is all customary and we refer the reader to any textbook in machine learning for details. 

The choice of loss $\ell$ is specific to the task of interest. For multi-class classification, it is customary to use (average) empirical cross-entropy, represented using the assumptions outlined above, as
\begin{equation}
\ell(x_i, y_i) = -\log \phi_w(x_i)[y_i].
\end{equation}
The minimizer of (average) empirical cross-entropy can be shown to be the minimizer of the average probability of error in a standard zero-one loss where every error has equal cost. This is not the case in medical image interpretation.

Crucial to Medical Imaging is the strong asymmetry between type-one errors (false alarms) that can result in unnecessary treatment and increased cost of care, and type-two errors (missed detection) that can be fatal. This must be taken into account in the computation of the average loss, or {\em risk}, which has to weigh each error by its cost. Another asymmetry is due to the incidence of disease: Because pathology are thankfully rare among the set of all studies conducted, a trivial classifier that always declares absence of diseases would achieve seemingly reasonable error rates. Of course, chance level is defined relative to the standard incidence of disease, and this is again a point of departure for Medical Imaging compared to standard natural image classification in the context of image search or content-based retrieval. Finally, the $K$ classes we consider are not necessarily mutually exclusive, and in some cases there are strong dependencies, so one being manifest affects the probability of others being too.

This is taken into account in DeepRadiologyNet by employing a knowledge graph that incorporates domain expertise from professional radiologists, using a hierarchical loss that penalizes different classes differently and accounts for lack of mutual exclusivity, and various balancing techniques to account for the prior distribution of diseases expected in the population.

In our hierarchical loss function,  pathologies are grouped according to multiple criteria, including pathology location, clinical significance and pathology type. A simple example groups pathologies based solely on their clinical significance; one possible grouping could be zero patient-risk, moderate patient risk and immediate with high risk to the patient.  In this scenario, should an image have a low or moderate risk label as well as a high risk label, the lower risk label is ignored in favor of the potentially life-threatening pathology. Another possible loss hierarchically groups pathologies based on their type and/or location. For example, intracranial hemorrage (ICH) could be one such grouping which is composed of pathologies like epidural hematoma, subdural hematoma, subarachnoid hemorrhage, intraventricular hemorrhage and parenchymal hemorrhage. 

A considerable amount of effort in the design of DeepRadiologyNet is, in addition to the choice of architectures and learning machinery described in the next section, in the curation of data and their management to ensure that population and disease priors are taken into account when specifying the composition of specialist networks in DeepRadiologyNet in a statistically correct manner, while satisfying known dependencies from domain expertise.

\begin{figure}[htp]
\centering
\includegraphics[width=.95\textwidth]{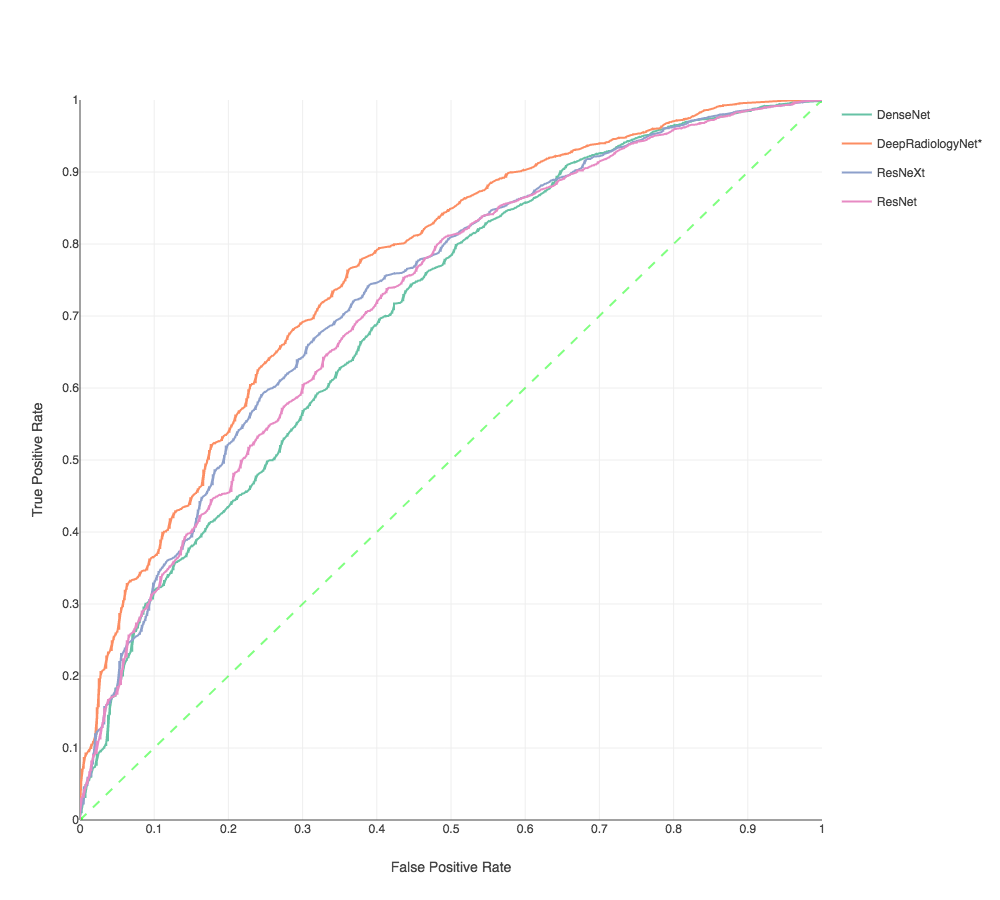}
\caption{Receiver operating curves measuring performance on detection of clinically significant traits of popular architectures \cite{huang2017densely,He2015,Xie2016} and the architecture used in DeepRadiologyNet. Validation was carried out on a set of 9000 studies.}\label{fig:net-comp-roc}
\end{figure}

\begin{figure}[htp]
\centering
\includegraphics[width=.95\textwidth]{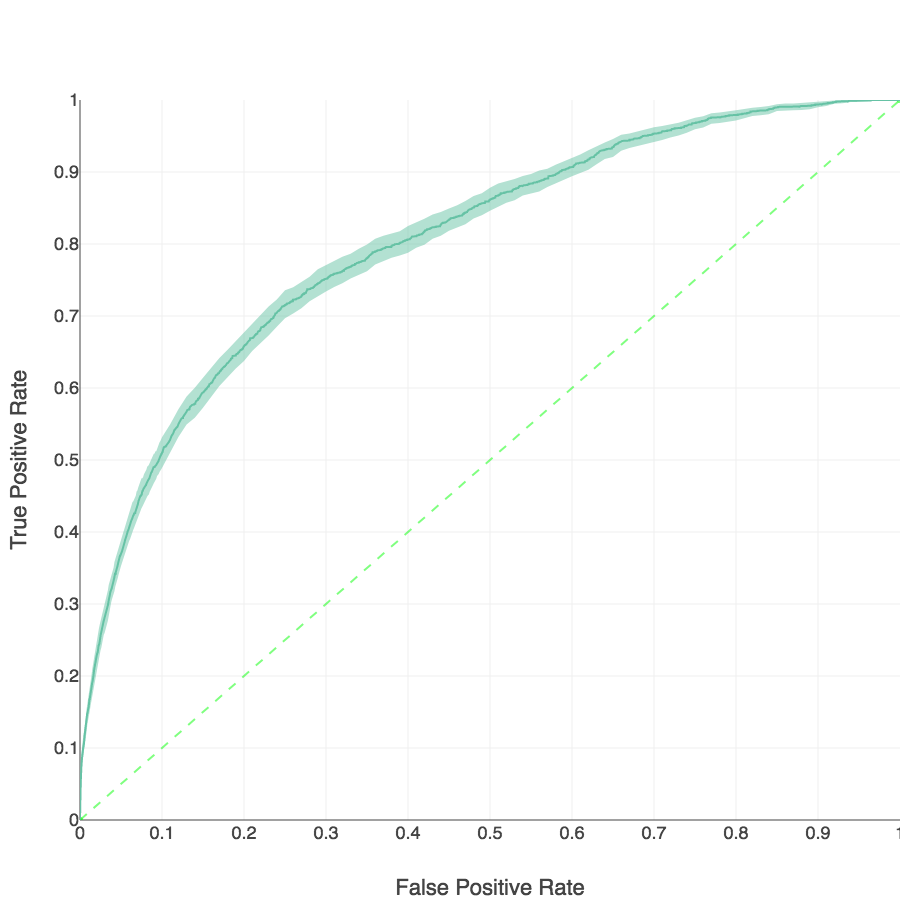}
\caption{Receiver operating curve measuring performance of DeepRadiogyNet on detection of clinically significant pathologies on the trial dataset of 29,965 studies, comprising over 4.8 million images.  95\% Confidence intervals are displayed as ribbon overlays, which were computed through bootstrap re-sampling of the data.}\label{fig:crit-roc}
\end{figure}

 \begin{figure}[htp]
 \centering
\includegraphics[width=.95\textwidth]{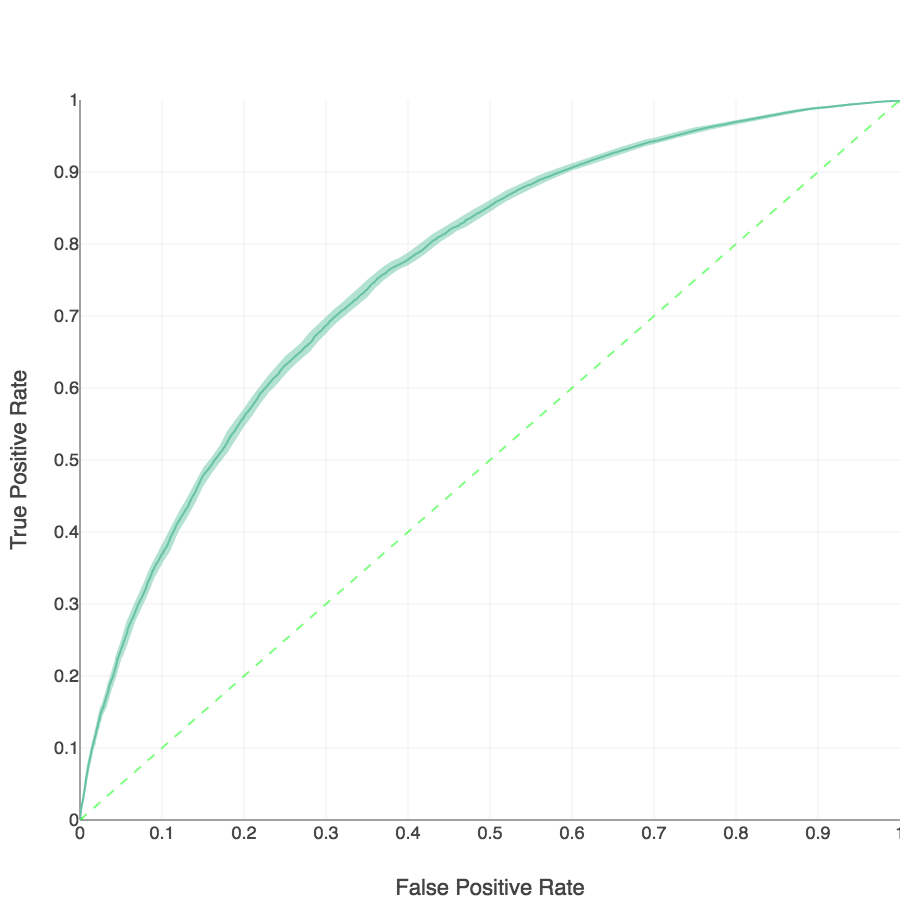}
\caption{Receiver operating curve measuring performance of DeepRadiogyNet on detection of {30} traits on the trial dataset of 29,965 studies, comprising over 4.8 million images.  95\% Confidence intervals are displayed as ribbon overlays, which were computed through bootstrap re-sampling of the data.}
\label{fig:path-roc}
\end{figure}

\section{Methods}

\begin{figure}[htp]
\begin{subfigure}{.5\textwidth}
\includegraphics[width=\textwidth]{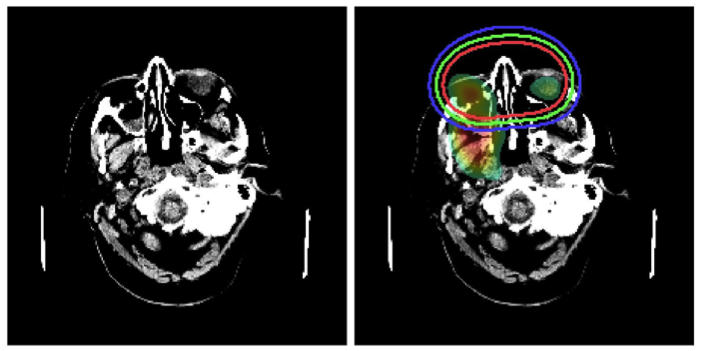}
\caption{\label{fig:saliency1}}
\end{subfigure}
\begin{subfigure}{.5\textwidth}
\includegraphics[width=\textwidth]{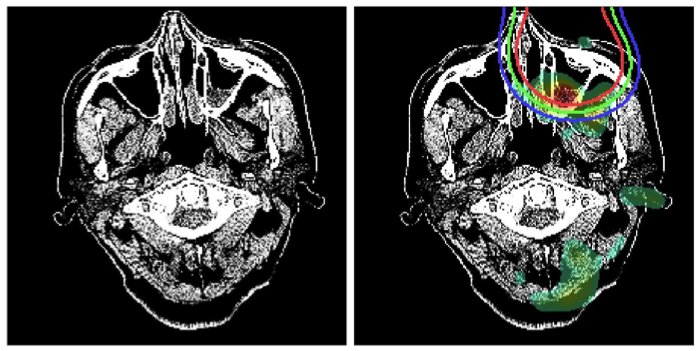}
\caption{\label{fig:saliency2}}
\end{subfigure}

\begin{subfigure}{.5\textwidth}
\includegraphics[width=\textwidth]{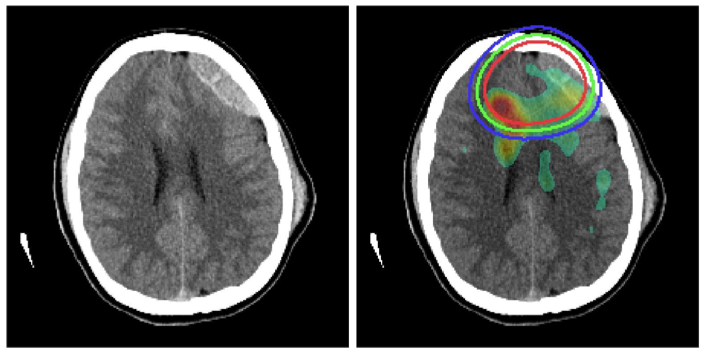}
\caption{\label{fig:saliency3}}
\end{subfigure}
\begin{subfigure}{.5\textwidth}
\includegraphics[width=\textwidth]{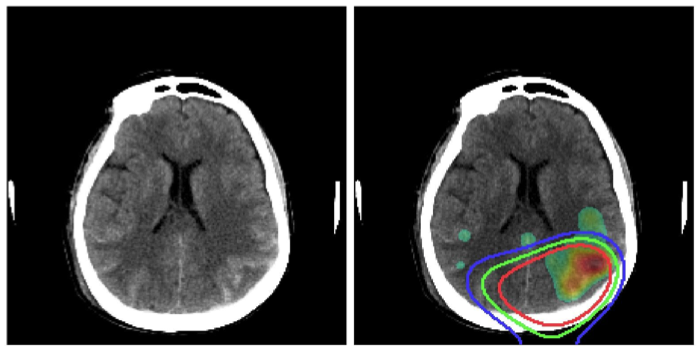}
\caption{\label{fig:saliency4}}
\end{subfigure}

\caption{Example visualization using our proprietary network introspection methodology. 
% \subref{fig:saliency1}, \subref{fig:saliency2}, \subref{fig:saliency3}\subref{fig:saliency4}
}
\label{fig:saliency}
\end{figure}

In the next section we describe the choice of architecture, loss function, and optimization, and in the following one we describe the methodology for data curation and evaluation.

\subsection{Technical description of DeepRadiologyNet}

Our network is composed of an ensemble of multiple GoogleNet-like networks \cite{googlenet}.
Figure \ref{fig:net-comp-roc} shows a comparison between the architecture used in DeepRadiolgyNet and other architecture choices: DenseNet \citep{huang2017densely}, ResNet \citep{He2015} and ResNeXt \citep{Xie2016}. These networks were trained on the same data as DeepRadiologyNet, and the comparison was carried out on separate validation set of over 9000 studies.

Each network in DeepRadiologyNet was trained starting from a different random initialization and traverse the training data in a different randomized order. DeepRadiologyNet contains networks that were trained using various base loss types including hinge loss and soft-max with multinomial logistic loss and may use different hierarchal loss mappings. Training updates were carried out using Adam optimization \cite{adam} or stochastic gradient descent with momentum. Training images were augmented, on-the-fly, by randomly rotating, rescaling, and mirroring images by clinically appropriate values. Networks were regularized through varying degrees of dropout and early stopping while training for a fixed number epochs with a \textit{step} learning rate policy.

\subsection{Data Annotation}
Training images were annotated by expert radiologists with a taxonomy of labels specifically developed for our deep learning processes including hierarchical loss.   Our taxonomies adhere to requirements for deep learning and AI, but also have clearly defined medical interpretations.  Given the size and scope of the data available, a method to locate studies which fit our annotation requirement was devised.  First, our taxonomies were mapped to specific keywords and phrases found in radiologist reports. These mappings allowed us to quickly find studies for our training set and prioritize image annotation by expert radiologists.

\subsection{Validation and Evaluation Methodology}

We measure the performance of DeepRadiologyNet with multiple metrics and labels, recording clinically insignificant errors, clinically significant errors, and validation diagnostics designed to be interpretable by human experts.

We produce receiver operator characteristics (ROC) curves and classifier error rates to enable modulating detection performance with functional usefulness: It is easy to achieve high precision at the expense of low recall and vice-versa. Carefully modulating the operating point is critical for the viability of an automated or assisted interpretation service.

The choice of operating point depends on acceptable levels of certainty which is highly pathology-dependent, application-dependent, and a complex issue not discussed in this manuscript. Here, we limit to showing our performance curves, to give the reader full access to the trade space. Confidence intervals (95\%) of our ROC curves are computed through bootstrap re-sampling.

\subsection{Radiologist Error Rates from Literature}
The types of discordance/disagreement of reports between radiologist is generally divided into two groups based on clinical impact: Clinically significant and clinically insignificant discordance \citep{erly2003evaluation}. We arrive at a literary error rate by collecting results from five sources which meet the criteria of reporting specific error rates in interpretation of CT head examinations by board certified radiologists.  \cite{pysher1999teleradiology} noted a 4\% clinically significant error rate in 137 CT scans of the head.   \cite{erly2003evaluation} found that in the reporting of 716 CT scans of the head, there was a clinically significant error rate of 2\% .   \cite{jordan2006quality} found a clinically significant error rate of 0.4\% in the interpretation of 1081 CT scans of the head.   Jordan et al. in 2012 reported that in 560 reports of CT head, there was a 0.7\% error rate \citep{jordan2012quality}. \cite{babiarz2012quality} reported a significant error rate of 2\% for 284 CT scans of the head. From the results of these five works, we calculate the overall error rate for the combined 2,778 CT head examinations to be 1.21\% through weighted summation. 

From this overall error rate, we wish to find a suitable clinically significant miss rate. \cite{renfrew1992error} reported that 81\% of errors where misses.   When looking just at CT head interpretations, \cite{erly2002radiology} noted a miss rate of 70\%, however, this study involved residents in training.  In a later work, \cite{erly2003evaluation} used practicing board certified radiologists with head CT interpretations found a false negative rate of 68\%.
We use the most conservative of these rates, and given the overall error rate of 1.21\% for CT head interpretation, of which  68\% are clinically significant misses, arrive at a {\em clinically significant miss rate (CSMR) of 0.83\%}, which we use for comparison to human performance in this manuscript. 

 \section{Clinical Trial}
In order to benchmark DeepRadiologyNet against CSMR by humans, we designed a retrospective clinical trial, conducted in 2017, using a set of studies completely disjoint from the training and validation set.
 
DeepRadiologyNet was trained on over 24,000 studies, containing approximately 3.5 million images and our trial was performed on 29,965 studies, comprising of 4.8 million images. All medical data was stripped of any identifying information, stored and transmitted through HIPAA compliant protocols and devices.  Trial studies originated from over 80 sites across the globe during two continuous time periods: September  2015 through December 2015 and May 2016 through September 2017. Imaging data was collected from over 50 types of scanners from all major manufacturers and includes patients in all age groups from newborns and infants to geriatrics patients. Incoming data was pre-processed based on their DICOM data, ensuring that they have valid headers and pixel data. Any data which contained corrupted DICOM headers were excluded from the evaluation.   We use the meta-data in the DICOM header to select axial images. These are submitted to DeepRadiologyNet and scores were generated. The label set produced mirrors the hierarchical loss used for training, only a subset of which is ultimately used to render the final decision, depending on which clinical test is being conducted.
 \newcolumntype{x}[1]{>{\centering\let\newline\\\arraybackslash\hspace{0pt}}m{#1}}
 \newcommand{\specialcell}[2][c]{%
 \renewcommand{\arraystretch}{0.8}%
  \begin{tabular}[#1]{@{}c@{}}#2\end{tabular}}
 
\afterpage{\clearpage
\begin{longtable}{m{1.5em}m{15em}m{3em}m{5em}m{5em}}
\caption{Population density of clinical trial studies and those fully characterized by DeepRadiologyNet.
The first column is the list of pathology ground-truth detected through multiple validation by human specialists (between 2 and 5 board-certified head radiologists), as well as clinical follow-through and outcomes. The second column is the total incidence of this pathology in the test set. The following two columns are the percentage errors reported by a network with operating point chosen to report on 42.1\% of the cases, automatically determine by the network based on a confidence score generated at test-time, and the same for a network reporting on 8.5\% of the cases.
The last rows indicate clinically significant errors, as defined and described in the text.
% \textcolor{red}{[Verbose trait names need to be verified by Rob+Kim.]}
}\label{tab:table2}\\
{} &          & Trial  & Net-42.1 & Net-8.5 \\
\midrule
\midrule
\endfirsthead
\multicolumn{5}{c}%
        {{\bfseries Table \thetable\ Continued from previous page}} \\
{Trait} &          & Trial  & Net-42.1 & Net-8.5 \\
\midrule
\midrule
\endhead      
{} & Artifact/Metal &            0.184 &       0.127 &     0.0395 \\
{} & Aneurysm &            0.418 &       0.191 &      0.079 \\
{} & Arachnoid Cyst &            0.468 &       0.366 &      0.276 \\
{} & Atherosclerosis &             1.01 &       0.302 &     0.0395 \\
{} & Calcification &             3.74 &        2.15 &       1.15 \\
{} & Cerebral Edema &            0.241 &      0.0318 &          0 \\
{} & Colloid Cyst&           0.0636 &      0.0397 &          0 \\
{} & Diffuse Volume Loss &             16.3 &        4.78 &       0.79 \\
{} & Encephalomalacia &             2.24 &       0.517 &      0.118 \\
{} & Fracture &             2.93 &        1.44 &      0.553 \\
{} & Glioblastoma Multiforme &           0.0167 &           0 &          0 \\
{} & Gun Shot Wound &            0.104 &      0.0477 &          0 \\
{} & Hydrocephalus &            0.733 &       0.127 &      0.079 \\
{} & Lacune &             2.39 &        0.89 &      0.197 \\
{} & Lipoma &             0.13 &       0.103 &     0.0395 \\
{} & Mastoid Pathology &             1.48 &       0.779 &      0.513 \\
{} & Meningioma &            0.475 &       0.223 &      0.079 \\
{} & Metastasis &            0.284 &       0.135 &     0.0395 \\
{} & Midline Shift &             1.45 &      0.0795 &     0.0395 \\
{} & Neoplasm &             0.14 &      0.0318 &          0 \\
{} & Chronic/Old Infarction &            0.518 &       0.183 &          0 \\
{} & Orbital Pathology &             2.03 &        1.13 &      0.395 \\
{} & Pneumocephalus &            0.572 &      0.0556 &          0 \\
{} & Scalp or Soft Tissue Pathology &             10.2 &        5.69 &       2.96 \\
{} & Sinus Pathology &             16.5 &        9.63 &       3.99 \\
{} & White Matter Ischemia &               17 &         4.4 &      0.553 \\
\midrule
{} & &                &         &   \\
\parbox[t]{3mm}{
% 	  \hspace{-.5em}
	  \multirow{4}{*}{
% 	  \hspace{-.5em}
	  \rotatebox[origin=c]{90}{
	  \specialcell{\scriptsize Clinically\\ \scriptsize Significant}}}}
{} & Acute Infarction &            0.241 &      0.0954 &          0 \\
{} & Depressed Skull Fracture &           0.0167 &     0.00795 &          0 \\
{} & Intracranial Hemorrhage &             5.14 &        1.24 &      0.316 \\
{} & Intracranial Mass &              2.2 &       0.564 &      0.118 \\
\bottomrule
\end{longtable}%
}

\begin{table}[ht]
 \centering
 \caption{Clinically significant miss rates of DeepRadiologyNet and radiologist.}
 \label{tab:table1}
 \begin{tabular}{m{15em}m{8em}}
Source & CSMR \\
\midrule
Literary   & 0.83\% \\
DeepRadiologyNet-42.1 & 0.802\% \\
DeepRadiologyNet-8.5 & 0.037\% \\
\bottomrule
\end{tabular}
 \end{table}
The analysis is conducted at the level of a study, rather than individual image, and the scores of individual images are aggregated through the study based on the uncertainty estimate of the label distribution produced by the network, $\phi_{\hat w}(x)$, interpreted as a posterior score relative to the distribution of labels in the hierarchy.

Studies in our clinical trial were exhaustively annotated with 30 non-mutually exclusive pathological traits which were divided based on their clinical significance.  Examples of clinically less significant traits include paranasal sinus disease, scalp swelling, old infarcts,  and chronic age related findings \citep{wong2005outsourced,erly2003evaluation}. 
Significant traits include those that could affect immediate management or have an adverse patient outcome such as acute intracranial hemorrhage, depressed skull fracture, acute infarction or intracranial mass. In our analysis, we look at performance in predicting all types of traits, however we pay careful attention to clinically significant findings.
The distribution of phenomenological traits in our clinical trial is shown in Figure \ref{fig:chord}.

For our clinical trial, we calculated clinically significant misses of all 29,965 studies through outcome analysis and consensus {of 2 and up to 5} radiologists for DeepRadiologyNet. 
DeepRadiologyNet had sufficient confidence to characterize and predict all {30} traits in 8.5\% of studies in the trial with a CSMR of 0.037\%, a rate that is {far below the literal estimated CSMR of board certified radiologist derived in the previous section.} Furthermore, at a different operating point, DeepRadiologyNet had confidence to report on 42.1\% of the studies with a lower CSMR than the estimated rate from literature.
These findings are summarized in \ref{tab:table1}.
Population characterization of the clinical trial data and studies which our DeepRadiologyNet reported on are summarized in Table \ref{tab:table2}.
Example localization of these predictions is depicted in Figure \ref{fig:saliency}. Localization was 
performed through a proprietary method beyond the scope of this manuscript.

These are just two sample operating points of DeepRadiologyNet. We envision that the choice of operating point will need to take into account a variety of factors including disease, modality, health-care provider, availability of human resources, geography and access to facilities, among other considerations.

\section{Discussion}
We have described a system and method to perform automated diagnosis of pathologies from CT head, developed over the course of multiple years and first disclosed at RSNA in December 2016. Since then, other studies have been reported in literature describing the use of deep learning, and specifically deep convolutional networks, for medical imaging interpretation. For the most part, these studies are too small to positively assess the clinical viability of deep learning as an automated diagnostic tool. Its role in both automated as well as assisted diagnostics remain to be fully validated. Even at comparable-to-human error rates, there may be advantages in deploying an automated system in a ``second-opinion mode,'' provided its operation does not bias the work of the radiologist, since the failure modes of the two are complementary. There is ample evidence that most errors made by trained and board-certified radiologists is due to inattention, whereas automated diagnostics may miss rare pathologies for which insufficient training data is available, but they do not get distracted. 

Comparing to human radiologists is non-trivial. First, humans define ground truth, so it is necessary to have multiple readings, as we do. 
Second, we are not choosing the ``best'' subset on which to report, but rather the network itself selects \emph{automatically} the studies to report on.
Choosing the ``best'' subset requires ground truth to be determined, which the network does not have access to. Instead, it performs a selection based on its own confidence, which is determined automatically without access to the ground truth. It is then meaningful to compute errors in the confidence subset only because the non-confident cases are referred to a radiologist. Note that false positives are zero by construction as the system only reports on high confidence negatives.

DeepRadiologyNet is continuously being improved and updated, and we believe its deployment will result in better care: faster, more accurate than humans, and far more cost-effective. More importantly, its continuous improvement and deployable nature make for high-quality diagnosis available in remote or under-served regions, or easing the bottleneck due to the shortage of highly trained specialists.

\section*{Acknowledgments}

This manuscript includes work conducted in 2015 and 2016 while S. Soatto, Z. Tu, A. Vedaldi were supported in part by DeepRadiology, INC. Disclosure under US 62/275,064; January 5, 2016.

\clearpage
\bibliographystyle{elsarticle-harv}
\bibliography{main}

\end{document}